\documentclass[10pt,twocolumn,letterpaper]{article}

\usepackage{wacv}
\usepackage{times}
\usepackage{epsfig}
\usepackage{graphicx}
\usepackage{amsmath}
\usepackage{amssymb}

\usepackage{booktabs}
\usepackage{multirow}
\usepackage{floatrow}
\usepackage{subcaption}
\usepackage{authblk}

\newcommand\CoAuthorMark{\footnotemark[\arabic{footnote}]}

\def\wacvPaperID{608} 

\wacvfinalcopy 

\ifwacvfinal
\fi

\ifwacvfinal
\usepackage[pagebackref=true,breaklinks=true,colorlinks,bookmarks=false]{hyperref}
\else
\usepackage[pagebackref=true,breaklinks=true,colorlinks,bookmarks=false]{hyperref}
\fi

\ifwacvfinal
\else
\fi

\begin{document}

\title{Temporal Context Aggregation for Video Retrieval with Contrastive Learning}

\author[1,3]{Jie Shao\thanks{Both authors contributed equally to this work.}}
\author[2,3]{Xin Wen\protect\CoAuthorMark\thanks{Work done while Xin Wen was an intern at ByteDance AI Lab.}}
\author[2]{Bingchen Zhao}
\author[1]{Xiangyang Xue}
\affil[1]{School of Computer Science, Fudan University, Shanghai, China}
\affil[2]{Department of Computer Science and Technology, Tongji University, Shanghai, China}
\affil[3]{ByteDance AI Lab}
\affil[ ]{\tt\small shaojie@fudan.edu.cn, wx99@tongji.edu.cn, zhaobc.gm@gmail.com, xyxue@fudan.edu.cn}

\maketitle

\begin{abstract}
The current research focus on Content-Based Video Retrieval requires higher-level video representation describing the long-range semantic dependencies of relevant incidents, events, etc. However, existing methods commonly process the frames of a video as individual images or short clips, making the modeling of long-range semantic dependencies difficult. In this paper, we propose TCA (Temporal Context Aggregation for Video Retrieval), a video representation learning framework that incorporates long-range temporal information between frame-level features using the self-attention mechanism. To train it on video retrieval datasets, we propose a supervised contrastive learning method that performs automatic hard negative mining and utilizes the memory bank mechanism to increase the capacity of negative samples. Extensive experiments are conducted on multiple video retrieval tasks, such as CC\_WEB\_VIDEO, FIVR-200K, and EVVE. The proposed method shows a significant performance advantage ($\sim17\%$ mAP on FIVR-200K) over state-of-the-art methods with video-level features, and deliver competitive results with 22x faster inference time comparing with frame-level features.
\end{abstract}

\section{Introduction}
We address the task of Content-Based Video Retrieval. The research focus on Content-Based Video Retrieval has shifted from Near-Duplicate Video Retrieval (NDVR)~\cite{wu2007practical,jiang2014vcdb} to Fine-grained Incident Video Retrieval~\cite{kordopatis2019fivr}, Event-based Video Retrieval~\cite{revaud2013event}, \etc. Different from NDVR, these tasks are more challenging in terms that they require higher-level representation describing the long-range semantic dependencies of relevant incidents, events, \etc.

\begin{figure}[t]
    \centering
    \includegraphics[width=\textwidth]{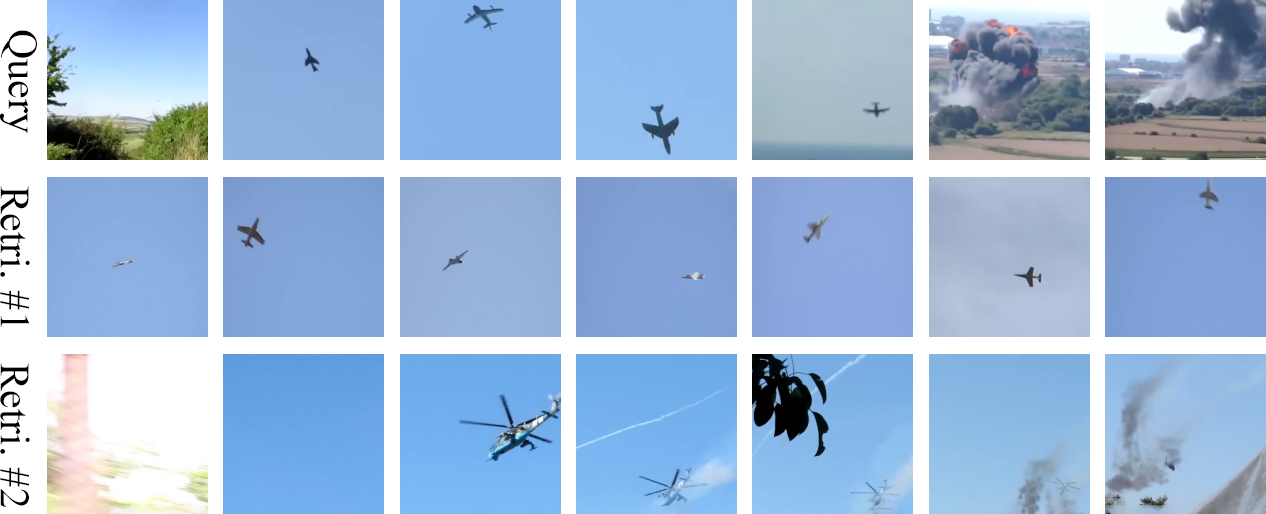}
    \caption{Example query describing the crash of a hawker hunter at Shoreham airport and its challenging distractors retrieved from the FIVR-200K~\cite{kordopatis2019fivr} dataset. As the scene of the aircraft in the sky takes the majority of the video, the vital information about the crash (with fewer frames) is covered up, thus the mistakenly retrieved videos share similar scenes, but describe totally different events.}
    \label{fig:badcase}
\end{figure}

The central task of Contend-Based Video Retrieval is to predict the similarity between video pairs. Current approaches mainly follow two schemes: to compute the similarity using video-level representations (first scheme) or frame-level representations (second scheme).
For methods using video-level representations, early studies typically employ code books~\cite{cai2011million,kordopatis2017near,liao2018ir} or hashing functions~\cite{song2011multiple,song2013effective} to form video representations, while later approach (Deep Metric Learning~\cite{kordopatis2017dml}) is introduced to generate video representations by aggregating the pre-extracted frame-level representations. In contrast, the approaches following the second scheme typically extract frame-level representations to compute frame-to-frame similarities, which are then used to obtain video-level similarities~\cite{chou2015pattern,Liu2017An,kordopatis2019visil,tan2009scalable}. 
With more elaborate similarity measurements, they typically outperform those methods with the first scheme.

For both schemes, the frames of a video are commonly processed as individual images or short clips, making the modeling of long-range semantic dependencies difficult. As the visual scene of videos can be redundant (such as scenery shots or B-rolls), potentially unnecessary visual data may dominate the video representation, and mislead the model to retrieve negative samples sharing similar scenes, as the example shown in Fig.~\ref{fig:badcase}. 
Motivated by the effectiveness of the self-attention mechanism in capturing long-range dependencies~\cite{vaswani2017attention}, we propose to incorporate temporal information between frame-level features (\ie, temporal context aggregation) using the self-attention mechanism to better model the long-range semantic dependencies, helping the model focus on more informative frames, thus obtaining more relevant and robust features.

To supervise the optimization of video retrieval models, current state-of-the-art methods~\cite{kordopatis2017dml,kordopatis2019visil} commonly perform pair-wise optimization with triplet loss~\cite{weinberger2009distance}. However, the relation that triplets can cover is limited, and the performance of triplet loss is highly subject to the time-consuming hard-negative sampling process~\cite{sohn2016improved}.
Inspired by the recent success of contrastive learning on self-supervised learning~\cite{he2019momentum,chen2020simple} and the nature of video retrieval datasets that rich negative samples are readily available, we propose a supervised contrastive learning method for video retrieval. With the help of a shared memory bank, large quantities of negative samples are utilized efficiently with no need for manual hard-negative sampling.
Furthermore, by conducting gradient analysis, we show that our proposed method has the property of automatic hard-negative mining which could greatly improve the final performance.

\begin{figure}[t]
    \centering
    \includegraphics[width=\textwidth]{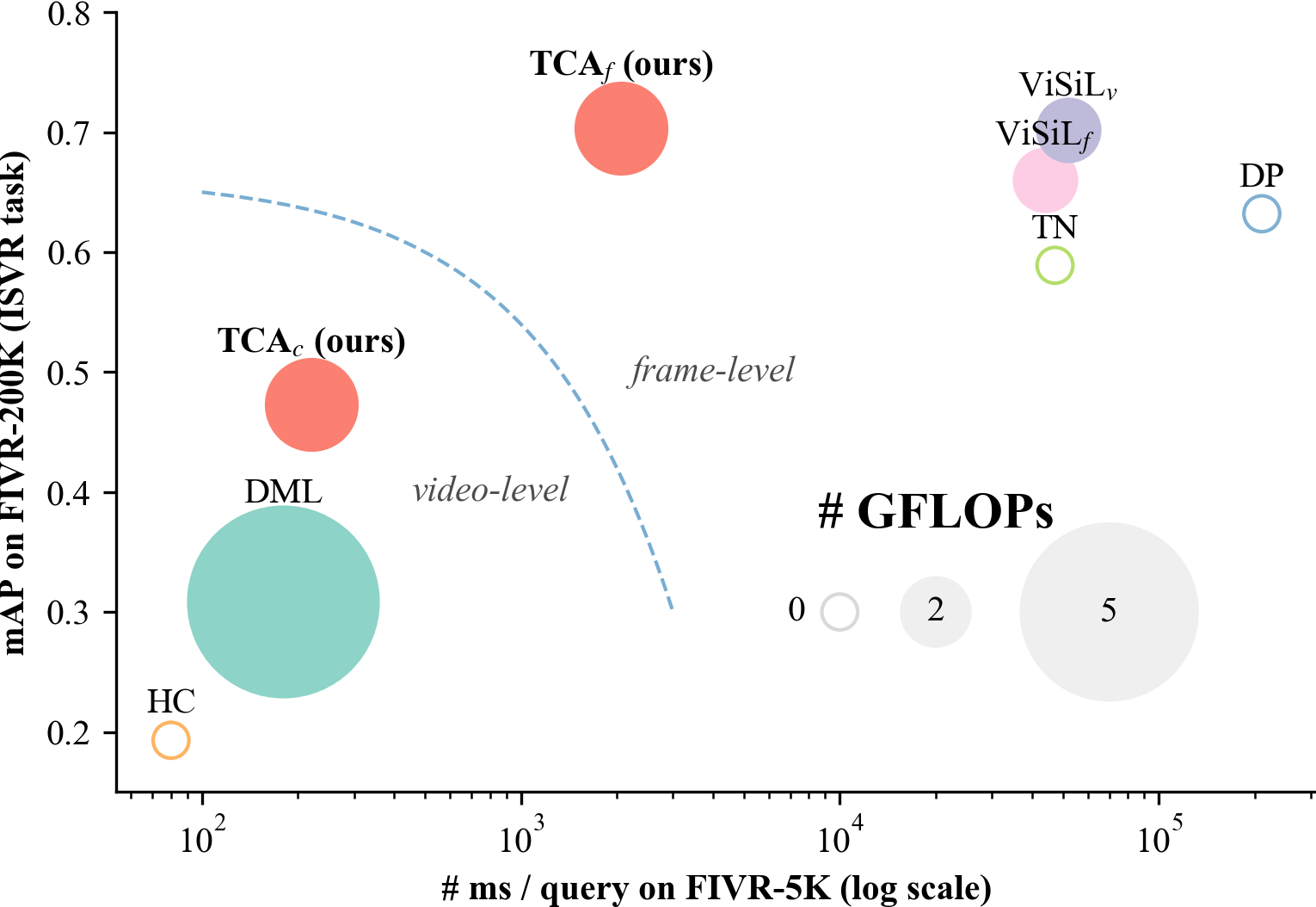}
    \caption{Video Retrieval performance comparison on ISVR task of FIVR~\cite{kordopatis2019fivr} in terms of mAP, inference time, and computational cost of the model (ISVR is the most complete and hard task of FIVR). The proposed approach achieves the best trade-off between performance and efficiency with both video-level and frame-level features against state-of-the-art methods. (\textit{Best viewed in color})}
    \label{fig:perf}
\end{figure}

Extensive experiments are conducted on multi video retrieval datasets, such as CC\_WEB\_VIDEO~\cite{wu2007practical}, FIVR~\cite{kordopatis2019fivr}, and EVVE~\cite{revaud2013event}.
In comparison with previous methods, as shown in Fig.~\ref{fig:perf}, the proposed method shows a significant performance advantage (\eg, $\sim17\%$ mAP on FIVR-200K) over state-of-the-art methods with video-level features, and deliver competitive results with 22x faster inference time comparing with methods using frame-level features.

\section{Related Work}

\textbf{Frame Feature Representation.}
Early approaches employed handcrafted features including the Scale-Invariant Feature Transform (SIFT) features ~\cite{jiang2007towards, lowe2004distinctive, wu2007practical}, the Speeded-Up Robust Features (SURF) ~\cite{bay2006surf, chou2015pattern}, Colour Histograms in HSV space ~\cite{hao2016stochastic, jing2019global, song2013effective}, and Local Binary Patterns (LBP) ~\cite{zhao2007dynamic, shang2010real, wu2014self}, \textit{etc.}
Recently, Deep Convolutional Neural Networks (CNNs) have proved to be versatile representation tools in recent approaches. The application of Maximum Activation of Convolutions (MAC) and its variants~\cite{razavian2016visual,zheng2016good,radenovic2016cnn,tolias2015particular,zheng2017sift,seddati2017towards,gordo2017end}, which extract frame descriptors from activations of a pre-trained CNN model, have achieved great success in both fine-grained image retrieval and video retrieval tasks~\cite{gordo2017end,kordopatis2017near,li2017ms,kordopatis2017dml,kordopatis2019visil}.
Besides variants of MAC, Sum-Pooled Convolutional features (SPoC)~\cite{babenko2015aggregating} and Generalized Mean (GeM)~\cite{hao2017unsupervised} pooling are also considerable counterparts.

\textbf{Video Feature Aggregation.}
Typically, the video feature aggregation paradigm can be divided into two categories: (1) local feature aggregation models~\cite{csurka2004visual,sivic2003video,perronnin2007fisher,jegou2010aggregating} which are derived from traditional local image feature aggregation models, and (2) sequence models ~\cite{hochreiter1997long,cho2014properties,donahue2015long,feng2018video,vaswani2017attention,Xia2019WeaklySE} that model the temporal order of the video representation.
Popular local feature aggregation models include Bag-of-Words~\cite{csurka2004visual,sivic2003video}, Fisher Vector~\cite{perronnin2007fisher}, and Vector of Locally Aggregated Descriptors (VLAD)~\cite{jegou2010aggregating}, of which the unsupervised learning of a visual code book is required. The NetVLAD~\cite{arandjelovic2016netvlad} transfers VLAD into a differential version, and the clusters are tuned via back-propagation instead of k-means clustering.
In terms of the sequence models, the Long Short-Term Memory (LSTM)~\cite{hochreiter1997long} and Gated Recurrent Unit (GRU)~\cite{cho2014properties} are commonly used for video re-localization and copy detection~\cite{feng2018video,hu2018learning}. Besides, self-attention mechanism also shows success in video classification~\cite{wang2018non} and object detection~\cite{hu2018relation}.

\textbf{Contrastive Learning.}
Contrastive learning has become the common training paradigm of recent self-supervised learning works~\cite{oord2018representation,hjelm2018learning,tian2019contrastive,he2019momentum,chen2020simple}, in which the positive and negative sample pairs are constructed with a pretext task in advance, and the model tries to distinguish the positive sample from massive randomly sampled negative samples in a classification manner. The contrastive loss typically performs better in general than triplet loss on representation learning~\cite{chen2020simple}, as the triplet loss can only handle one positive and negative at a time. The core of the effectiveness of contrastive learning is the use of rich negative samples~\cite{tian2019contrastive}, one approach is to sample them from a shared memory bank~\cite{wu2018unsupervised}, and~\cite{he2019momentum} replaced the bank with a queue and used a moving-averaged encoder to build a larger and consistent dictionary on-the-fly.

\section{Method} \label{section:method}
In this section, we ﬁrst deﬁne the problem setting (Section \ref{subsection:probset}) and describe the frame-level feature extraction step (Section \ref{subsection:featextract}). Then, we demonstrate the temporal context aggregation module (Section \ref{subsection:featagg}) and the contrastive learning method based on pair-wise video labels (Section \ref{subsection:contlearn}), then conduct further analysis on the gradients of the loss function (Section \ref{subsection:onestepfurther}). And last, we discuss the similarity measure of video-level and frame-level video descriptors (Section \ref{subsection:simcal}).

\subsection{Problem Setting} \label{subsection:probset}

We address the problem of video representation learning for Near-Duplicate Video Retrieval (NDVR), Fine-grained Incident Video Retrieval (FIVR), and Event Video Retrieval (EVR) tasks. In our setting, the dataset is two-split: the \textit{core} and \textit{distractor}. The core subset contains pair-wise labels describing which two videos are similar (near duplicate, complementary scene, same event, \etc). And the distractor subset contain large quantities of negative samples to make the retrieval task more challenging.

We only consider the RGB data of the videos. Given raw pixels ($\mathbf{x}_{r} \in \mathbb{R}^{m\times n\times f}$), a video is encoded into a sequence of frame-level descriptors ($\mathbf{x}_{f} \in \mathbb{R}^{d \times f}$) or a compact video-level descriptor ($\mathbf{x}_{v} \in \mathbb{R}^{d}$). Take the similarity function as $\text{sim}(\cdot,\cdot)$, the similarity of two video descriptors $\mathbf{x}_{1}, \mathbf{x}_{2}$ can be denoted as $\text{sim}(\mathbf{x}_{1}, \mathbf{x}_{2})$. Given these, our task is to optimize the embedding function $f(\cdot)$, such that $\text{sim}\left( f\left(\mathbf{x}_{1}\right), f\left(\mathbf{x}_{2}\right) \right)$ is maximized if $\mathbf{x}_{1}$ and $\mathbf{x}_{2}$ are similar videos, and minimized otherwise. The embedding function $f(\cdot)$ typically takes a video-level descriptor $\mathbf{x} \in \mathbb{R}^d$ and returns an embedding $f(\mathbf{x}) \in \mathbb{R}^k$, in which $k \ll d$. However, in our setting, $f(\cdot)$ is a 
temporal context aggregation modeling module, thus frame-level descriptors $\mathbf{x} \in \mathbb{R}^{d \times f}$ are taken as input, and the output can be either aggregated video-level descriptor ($f(\mathbf{x}) \in \mathbb{R}^{d}$) or refined frame-level descriptors ($f(\mathbf{x}) \in \mathbb{R}^{d \times f}$).

\subsection{Feature Extraction} \label{subsection:featextract}
According to the results reported in ~\cite{kordopatis2019visil} (Table 2), we select iMAC~\cite{gordo2017end} and modified $\text{L}_{3}$-iMAC~\cite{kordopatis2019visil} (called $\text{L}_3$-iRMAC) as our benchmark frame-level feature extraction methods. Given a pre-trained CNN network with $K$ convolutional layers, $K$ feature maps $\mathcal{M}^{k} \in \mathbb{R}^{n_{d}^{k}\times n_{d}^{k}\times c^{k}}(k=1,\dots,K)$ are generated, where
$n_{d}^{k}\times n_{d}^{k}$ is the dimension of each feature map of the $k^{\text{th}}$ layer, and $c^k$ is the total number of channels. 

For iMAC feature, the maximum value of every channel of each layer is extracted to generate $K$ feature maps $\mathcal{M}^{k} \in \mathbb{R}^{c^k}$, as formulated in Eq.~\ref{eq:imac}:
\begin{equation}
    \label{eq:imac}
    v^{k}(i) = \max \mathcal{M}^{k}(\cdot, \cdot, i),\quad i=1,2,\dots,c^{k} \,,
\end{equation}
where $v^{k}$ is a $c^{k}$-dimensional vector that is derived from max pooling on each channel of the feature map $\mathcal{M}^{k}$.

Max pooling with different kernel size and stride are applied to every channel of different layers to generate $K$ feature maps $\mathcal{M}^{k} \in \mathbb{R}^{3\times 3\times c^k}$ in the original $\text{L}_{3}$-iMAC feature. Unlike its setting, we then follow the tradition of R-MAC~\cite{tolias2015particular} to sum the $3\times3$ feature maps together, then apply $\ell_{2}$-normalization on each channel to form a feature map $\mathcal{M}^{k} \in \mathbb{R}^{c^k}$. This presents a trade-off between the preservation of fine-trained spatial information and low feature dimensionality (equal to iMAC), we denote this approach as $\text{L}_3$-iRMAC.

For both iMAC and $\text{L}_3$-iRMAC, all layer vectors are concatenated to a single descriptor after extraction, then PCA is applied to perform whitening and dimensionality reduction following the common practice~\cite{jegou2012negative,kordopatis2019visil}, finally $\ell_{2}$-normalization is applied on each channel, resulting in a compact frame-level descriptor $\mathbf{x} \in \mathbb{R}^{d \times f}$.

\begin{figure}[t]
    \centering
    \includegraphics[width=0.9\textwidth]{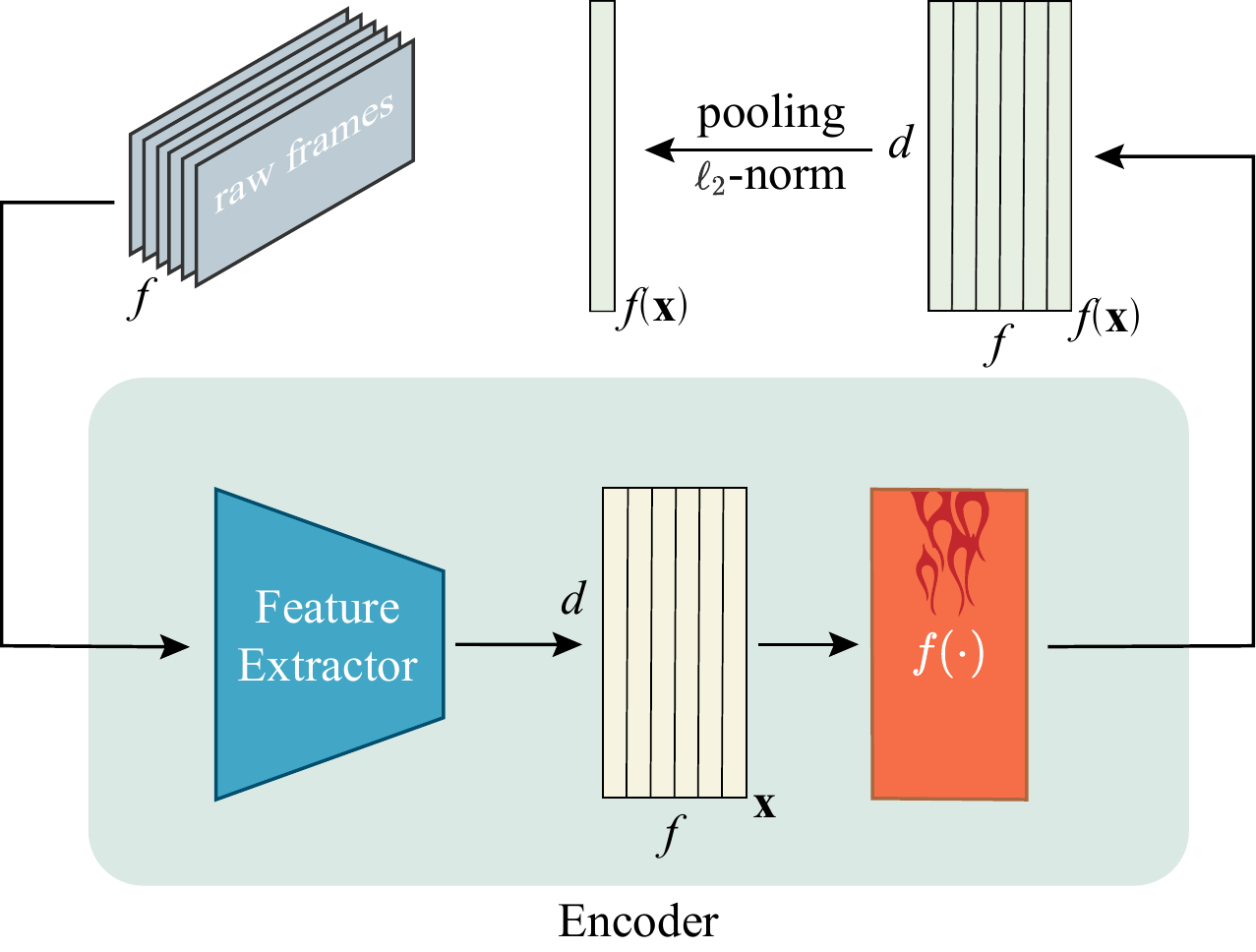}
    \caption{\textbf{Feature encoding pipeline.} Raw frames are fed to the feature extractor to extract the frame-level video descriptor $\mathbf{x}$. Then the self-attention mechanism is applied to perform temporal context aggregation on the input, and generate refined frame-level descriptors  $f(\mathbf{x})$. They can also be compressed into a video-level descriptor by applying average pooling and $\ell_2$-normalization.}
    \label{fig:feature}
\end{figure}

\subsection{Temporal Context Aggregation} \label{subsection:featagg}
We adopt the Transformer~\cite{vaswani2017attention} model for temporal context aggregation. Following the setting of~\cite{feng2018video,Xia2019WeaklySE}, only the encoder structure of the Transformer is used. With the parameter matrices written as $W^Q, W^K, W^V$, the entire video descriptor $\mathbf{x} \in \mathbb{R}^{d \times f}$ is first encoded into Query $Q$, Key $K$ and Value $V$ by three different linear transformations: $Q = \mathbf{x}^{\top}W^{Q}$, $K = \mathbf{x}^{\top}W^{K}$ and $V = \mathbf{x}^{\top}W^{V}$. This is further calculated by the self-attention layer as:
{\small
\begin{equation}
    \text{Attention}(Q,K,V) = \text{softmax}\left( \frac{QK^{\top}}{\sqrt{d}}\right)V \,.
\end{equation}
}
The result is then taken to the LayerNorm layer~\cite{ba2016layer} and Feed Forward Layer~\cite{vaswani2017attention} to get the output of the Transformer encoder, \ie, $f_{\text{Transformer}}(\mathbf{x}) \in \mathbb{R}^{d \times f}$. The multi-head attention mechanism is also used. 

With the help of the self-attention mechanism, Transformer is effective at modeling long-term dependencies within the frame sequence. Although the encoded feature keeps the same shape as the input, the contextual information within a longer range of each frame-level descriptor is incorporated. Apart from the frame-level descriptor, by simply averaging the encoded frame-level video descriptors along the time axis, we can also get the compact video-level representation $\overline{f}(\mathbf{x}) \in \mathbb{R}^{d}$. 

\subsection{Contrastive Learning} \label{subsection:contlearn}
If we denote $\mathbf{w}_a, \mathbf{w}_p, \mathbf{w}_n^j (j=1,2,\dots,N-1)$ as the video-level representation before applying normalization of the anchor, positive, negative examples, we get the similarity scores by: $s_p=\left. \mathbf{w}_{a}^{\top}\mathbf{w}_{p} \middle/ \left( \left\|\mathbf{w}_{a}\right\| \left\|\mathbf{w}_{p}\right\| \right) \right.$ and $s_n^j=\left. \mathbf{w}_{a}^{\top}\mathbf{w}_{n}^{j} \middle/ \left( \left\|\mathbf{w}_{a}\right\| \left\|\mathbf{w}_{n}^{j}\right\| \right) \right.$. Then the InfoNCE~\cite{oord2018representation} loss is written as:
{\small
\begin{equation} \label{eq:nce}
    \mathcal{L}_{\text{nce}}=-\log \frac{\exp \left(s_{p} / \tau\right)}{\exp \left(s_{p}\right) + \sum_{j=1}^{N-1} \exp \left( s_{n}^{j} / \tau\right)} \,,
\end{equation}
}
where $\tau$ is a temperature hyper-parameter~\cite{wu2018unsupervised}. To utilize more negative samples for better performance, we borrow the idea of the memory bank from~\cite{wu2018unsupervised}. For each batch, we take one positive pair from the core dataset and randomly sample $n$ negative samples from the distractors, then the compact video-level descriptors are generated with a shared encoder. The negative samples of all batches and all GPUs are concatenated together to form the memory bank. We compare the similarity of the anchor sample against the positive sample and all negatives in the memory bank, resulting in $1$ $s_p$ and $kn$ $s_n$. Then the loss can be calculated in a classification manner. The momentum mechanism~\cite{he2019momentum} is not adopted as we did not see any improvement in experiments. Besides the InfoNCE loss, the recent proposed Circle Loss~\cite{sun2020circle} is also considered:
{\small
\begin{equation} \label{eq:circle}
\mathcal{L}_{\text{circle}}= -\log \frac{\exp (\gamma \alpha_{p} ( s_{p} - \Delta_{p} ))}{\exp (\gamma \alpha_{p} ( s_{p} - \Delta_{p} )) + \sum\limits_{j=1}^{N-1} \exp (\gamma \alpha_{n}^{j} ( s_{n}^{j} - \Delta_{n} ))}
\end{equation}
}
where $\gamma$ is the scale factor(equivalent with the parameter $\tau$ in Eq.~\ref{eq:nce}), and $m$ is the relaxation margin. $\alpha_{p} = \left[ 1 + m - s_{p} \right]_{+}, \alpha_{n}^{j} = \left[ s_{n}^{j} + m\right]_{+}, \Delta_{p} = 1 - m, \Delta_{n} = m$. Compared with the InfoNCE loss, the Circle loss optimizes $s_p$ and $s_n$ separately with adaptive penalty strength and adds within-class and between-class margins. 

\begin{figure}[t]
    \centering
    \includegraphics[width=0.95\textwidth]{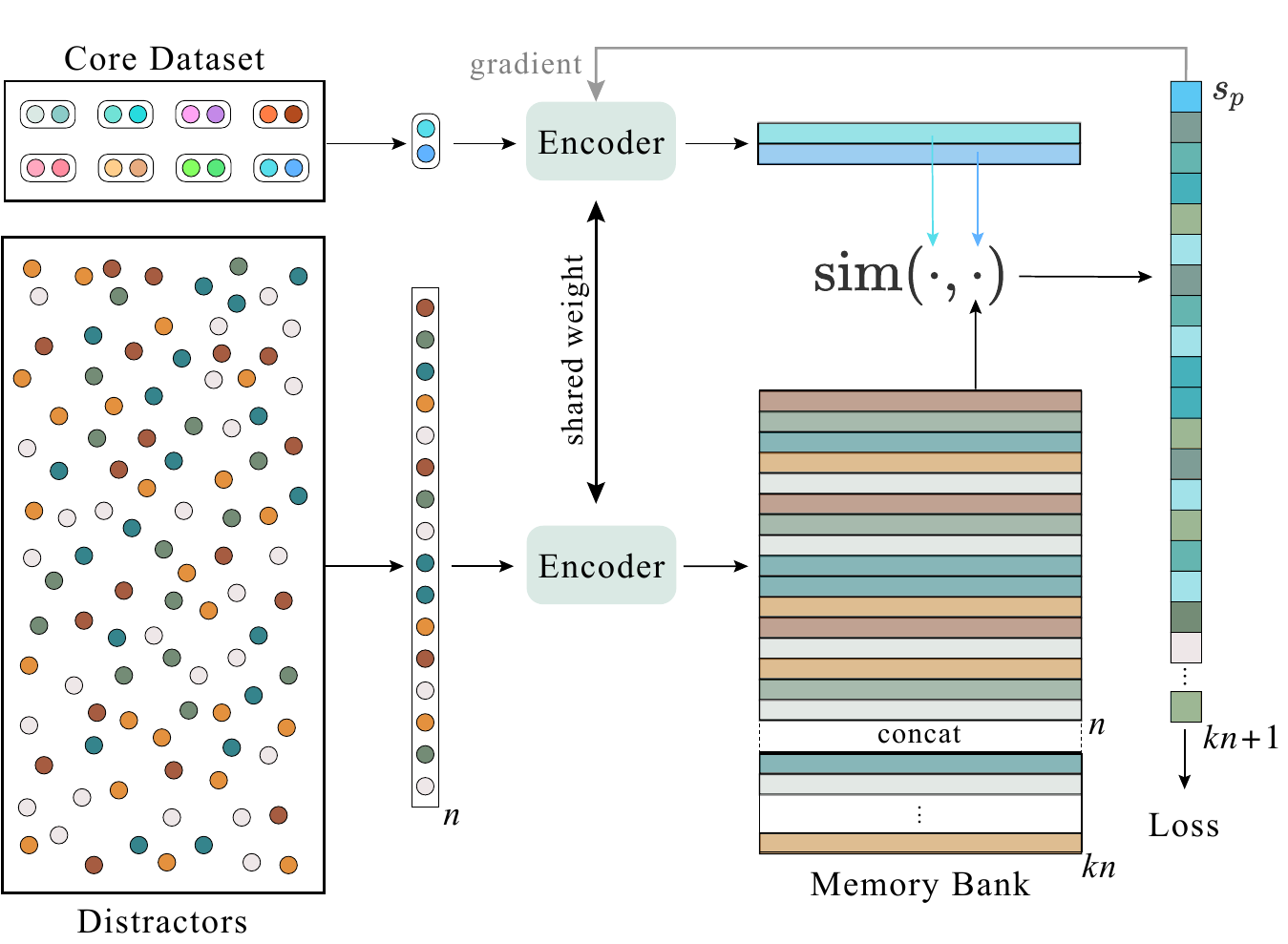}
    \caption{\textbf{Learning representation with pair-wise labels.} For each batch, we take one positive pair from the core dataset and randomly sample $n$ negative samples from the distractors, then the video-level descriptors are generated with a shared encoder. The negative samples of all batches and all GPUs are concatenated together to form the memory bank. We compare the similarity of the anchor against the positive sample and all negatives in the memory bank, resulting in $1$ $s_p$ and $kn$ $s_n$. Then the loss can be calculated in a classification manner following Eq.~\ref{eq:nce} and Eq.~\ref{eq:circle}.}
    \label{fig:contrast}
\end{figure}

\subsection{One Step Further on the Gradients} \label{subsection:onestepfurther}
In the recent work of Khosla \textit{et al.}~\cite{Khosla2020SupervisedCL}, the proposed batch contrastive loss is proved to focus on the hard positives and negatives automatically with the help of feature normalization by conducting gradient analysis, we further reveal that this is the common property of Softmax loss and its variants when combined with feature normalization. For simplicity, we analyze the gradients of Softmax loss, the origin of both InfoNCE loss and Circle loss:
{\small
\begin{equation} \label{eq:softmax}
    \mathcal{L}_{\text{softmax}}= -\log\frac{\exp\left(s_p\right)}{\exp\left(s_p\right)+\sum_{j=1}^{n-1}\exp\left(s_{n}^{j}\right)} \,,
\end{equation}
}
the notation is as aforementioned. Here we show that easy negatives contribute the gradient weakly while hard negatives contribute greater. With the notations declared in Section \ref{subsection:contlearn}, we denote the normalized video-level representation as $\mathbf{z}_{*} = \left. \mathbf{w}_{*} \middle/ \left\|\mathbf{w}_{*}\right\| \right.$, then the gradients of Eq.~\ref{eq:softmax} with respect to $\mathbf{w}_{a}$ is:
{\small
\begin{equation} \label{eq:softmaxw}
\begin{aligned}
    &\frac{\partial\mathcal{L}_{\text{softmax}}}{\partial \mathbf{w}_{a}}
    = \frac{\partial \mathbf{z}_{a}}{\partial \mathbf{w}_{a}} \cdot \frac{\partial \mathcal{L}_{\text{softmax}}}{\partial \mathbf{z}_{a}} \\
    &= \frac{1}{\left\| \mathbf{w}_{a} \right\|} \left( \mathbf{I} - \mathbf{z}_{a}\mathbf{z}_{a}^{\top} \right) \left[ \left( \sigma(\mathbf{s})_{p} - 1 \right)\mathbf{z}_{p} + \sum_{j=1}^{N-1}\sigma(\mathbf{s})_{n}^{j}\mathbf{z}_{n}^{j} \right] \\
    &\propto  
    \overbrace{( 1 - \sigma(\mathbf{s})_{p} )[ (\mathbf{z}_{a}^{\top}\mathbf{z}_{p})\mathbf{z}_{a} - \mathbf{z}_{p} ]}^{\text{positive}}
    +
    \underbrace{\sum_{j=1}^{N-1}\sigma(\mathbf{s})_{n}^{j}[ \mathbf{z}_{n}^{j} - (\mathbf{z}_{a}^{\top}\mathbf{z}_{n}^{j})\mathbf{z}_{a} ]}_{\text{negatives}} \,,
\end{aligned}
\end{equation}
}
where $\sigma(\mathbf{s})_{p} = \left. \exp\left(s_{p}\right) \middle/ \left[ \exp\left(s_{p}\right) + \sum_{j=1}^{N-1} \exp\left(s_{n}^{j}\right) \right] \right.$, and $\sigma(\mathbf{s})_{n}^{j} = \left. \exp\left(s_{n}^{j}\right) \middle/ \left[ \exp\left(s_{p}\right) + \sum_{j=1}^{N-1} \exp\left(s_{n}^{j}\right) \right] \right.$ following the common notation of the softmax function. For an easy negative, the similarity between it and the anchor is close to -1, thus $\mathbf{z}_{a}^{\top}\mathbf{z}_{n}^{j} \approx -1$, and therefore
{\small
\begin{equation} \label{eq:gradcontrib}
    \sigma(\mathbf{s})_{n}^{j} \left\| \left(\mathbf{z}_{n}^{j} -\left(\mathbf{z}_{a}^{\top}\mathbf{z}_{n}^{j}\right)\mathbf{z}_{a} \right) \right\| = \sigma(\mathbf{s})_{n}^{j} \sqrt{1 - \left( \mathbf{z}_{a}^{\top}\mathbf{z}_{n}^{j} \right)^2} \approx 0 \,.
\end{equation}
}
And for a hard negative, $\mathbf{z}_{a}^{\top}\mathbf{z}_{n}^{j} \approx 0$\footnote{This represents the majority of hard negatives, and if the similarity is close to 1, it is too hard and may cause the model to collapse, or due to wrong annotation.}, and $\sigma(\mathbf{s})_{n}^{j}$ is moderate, thus the above equation is greater than 0, and its contribution to the gradient of the loss function is greater. Former research only explained it intuitively that features with shorter amplitudes often represent categories that are more difficult to distinguish, and applying feature normalization would divide harder examples with a smaller value (the amplitude), thus getting relatively larger gradients~\cite{feng2018research}, however, we prove this property for the first time by conducting gradient analysis. The derivation process of Eq.~\ref{eq:nce} and Eq.~\ref{eq:circle} are alike. Comparing with the commonly used Triplet loss in video retrieval tasks~\cite{kordopatis2017dml,kordopatis2019visil} which requires computationally expensive hard negative mining, the proposed method based on contrastive learning takes advantage of the nature of softmax-based loss when combined with feature normalization to perform hard negative mining automatically, and use the memory bank mechanism to increase the capacity of negative samples, which greatly improves the training efficiency and effect.

\subsection{Similarity Measure} \label{subsection:simcal}
To save the computation and memory cost, at the training stage, all feature aggregation models are trained with the output as $\ell_2$-normalized video-level descriptors ($f(\mathbf{x}) \in \mathbb{R}^{d}$), thus the similarity between video pairs is simply calculated by dot product. Besides, for the sequence aggregation models, refined frame-level video descriptors ($f(\mathbf{x}) \in \mathbb{R}^{d\times f}$) can also be easily extracted before applying average pooling along the time axis. Following the setting in~\cite{kordopatis2019visil}, at the evaluation stage, we also use chamfer similarity to calculate the similarity between two frame-level video descriptors. Denote the representation of two videos as $\mathbf{x} = [x_0, x_1, \dots, x_{n-1}]^{\top}$, $\mathbf{y} = [y_0, y_1, \dots, y_{m-1}]^{\top}$, where $x_i, y_j \in \mathbb{R}^d$, the chamfer similarity between them is:
\begin{equation} \label{eq:chamfer}
    \text{sim}_{f}(\mathbf{x}, \mathbf{y}) = \frac{1}{n}\sum_{i=0}^{n-1}\max_{j}{x_{i}y_{j}^{\top}} \,,
\end{equation}
and the symmetric version:
\begin{equation} \label{eq:sym}
    \text{sim}_{sym}(\mathbf{x}, \mathbf{y}) = \left.\left( \text{sim}_{f}(\mathbf{x}, \mathbf{y}) + \text{sim}_{f}(\mathbf{y}, \mathbf{x}) \right) \middle/ 2\right. \,.
\end{equation}
Note that this approach (chamfer similarity) seems to be inconsistent with the training target (cosine similarity), where the frame-level video descriptors are averaged into a compact representation and the similarity is calculated with dot product. However, the similarity calculation process of the compact video descriptors can be written as:
{\small
\begin{equation} \label{eq:cos}
\begin{aligned}
    \text{sim}_{cos}(\mathbf{x}, \mathbf{y}) &= \left(\frac{1}{n}\sum_{i=0}^{n-1}x_{i}\right)\left(\frac{1}{m}\sum_{j=0}^{m-1}y_{j}\right)^{\top} \\
    &= \frac{1}{n}\sum_{i=0}^{n-1}\frac{1}{m}\sum_{j=0}^{m-1}x_{i}y_{j}^{\top} \,.
\end{aligned}
\end{equation}
}
Therefore, given frame-level features, chamfer similarity averages the maximum value of each row of the video-video similarity matrix, while cosine similarity averages the mean value of each row. It is obvious that $\text{sim}_{cos}(\mathbf{x}, \mathbf{y}) \leq \text{sim}_{f}(\mathbf{x}, \mathbf{y})$, therefore, by optimizing the cosine similarity, we are optimizing the lower-bound of the chamfer similarity. As only the compact video-level feature is required, both time and space complexity are greatly reduced as cosine similarity is much computational efficient.

\begin{table*}[t]
\floatbox{table}[\textwidth]{%
\begin{subfloatrow}
\floatbox{table}[.3\textwidth][\FBheight][t]{%
\caption{\textbf{Model} (mAP on FIVR-5K)} \label{subtab:model}}{
    \centering
    \setlength{\tabcolsep}{4pt}
    \begin{tabular}{lcccccc}
    Model & DSVR  & CSVR  & ISVR \\ 
    \toprule
    NetVLAD     & 0.513 & 0.494 & 0.412 \\
    LSTM        & 0.505 & 0.483 & 0.400 \\
    GRU         & 0.515 & 0.495 & 0.415 \\
    Transformer & \textbf{0.551} & \textbf{0.532} & \textbf{0.454} \\
    \end{tabular}
}
\floatbox{table}[.3\textwidth][\FBheight][t]{%
\caption{\textbf{Feature} (mAP on FIVR-200K)} \label{subtab:feature}}{
    \centering
    \setlength{\tabcolsep}{4pt}
    \begin{tabular}{lccc}
    Feature & DSVR  & CSVR  & ISVR  \\
    \toprule
    iMAC    & 0.547 & 0.526 & 0.447 \\
    $\text{L}_3$-iRMAC  & \textbf{0.570} & \textbf{0.553} & \textbf{0.473} \\
    \end{tabular}
}
\floatbox{table}[.4\textwidth][\FBheight][t]{%
\caption{\textbf{Loss function} (mAP on FIVR-200K)} \label{subtab:loss}}{
    \centering
    \setlength{\tabcolsep}{2pt}
    \begin{tabular}{lcccc}
    Loss & $\tau/\gamma$ & DSVR  & CSVR  & ISVR  \\
    \toprule
    InfoNCE & $0.07$   & 0.493          & 0.473          & 0.394          \\
    InfoNCE & $1/256$  & 0.566          & 0.548          & 0.468          \\
    Circle  & $256$    & \textbf{0.570} & \textbf{0.553} & \textbf{0.473} \\
    \end{tabular}
}
\end{subfloatrow}
\par\nointerlineskip\vspace{8pt}
\begin{subfloatrow}
\floatbox{table}[.3\textwidth][\FBheight][t]{%
\caption{\textbf{Bank size} (mAP on FIVR-5K)} \label{subtab:banksz}}{
    \centering
    \setlength{\tabcolsep}{2pt}
    \begin{tabular}{lcccc}
    Method  & Bank Size & DSVR              & CSVR              & ISVR   \\
    \toprule
    triplet & -         & 0.510             & 0.509             & 0.455 \\
    ours    & 256       & 0.605             & 0.615             & 0.575 \\    
    ours    & 4096      & 0.609             & \textbf{0.617}    & \textbf{0.578} \\
    ours    & 65536     & \textbf{0.611}    & \textbf{0.617}    & 0.574 \\ 
    \end{tabular}
}
\floatbox{table}[.3\textwidth][\FBheight][t]{%
\caption{\textbf{Momentum} (mAP on FIVR-5K)} \label{subtab:momentum}}{
    \centering
    \setlength{\tabcolsep}{2pt}
    \begin{tabular}{cccc}
    Momentum    & DSVR              & CSVR              & ISVR   \\
    \toprule
    0 (bank)    & \textbf{0.609}    & \textbf{0.617}    & \textbf{0.578} \\    
    0.1         & 0.606             & 0.612             & 0.569          \\
    0.9         & 0.605             & 0.611             & 0.568          \\ 
    0.99        & 0.602             & 0.606             & 0.561          \\ 
    0.999       & 0.581             & 0.577             & 0.520          \\
    \end{tabular}
}
\floatbox{table}[.36\textwidth][\FBheight][t]{%
\caption{\textbf{Similarity Measure} (mAP on FIVR-5K)} \label{subtab:simmeasure}}{
    \centering
    \setlength{\tabcolsep}{2pt}
    \begin{tabular}{lcccc}
    Similarity Measure  & DSVR              & CSVR              & ISVR   \\
    \toprule
    cosine              & 0.609          & 0.617          & 0.578          \\
    chamfer             & \textbf{0.844} & \textbf{0.834} & \textbf{0.763} \\
    symm. chamfer       & 0.763          & 0.766          & 0.711          \\
    chamfer+comparator  & 0.726          & 0.735          & 0.701          \\
    \end{tabular}
}
\end{subfloatrow}}
{\caption{\textbf{Ablations on FIVR about:} (a): Temporal context aggregation methods; (b): Frame feature representations; (c): Loss functions for contrastive learning ($\gamma=1/\tau$); (d) Size of the memory bank; (e) Momentum parameter of the queue of MoCo~\cite{he2019momentum}, degenerate to memory bank when set to 0; (f) Similarity measures (video-level and frame-level), comparator: the comparator network used in $\text{ViSiL}_{v}$~\cite{kordopatis2019visil}, with original parameters retained.} \label{tab:ablation}}
\end{table*}
\section{Experiments} \label{section:experiments}
\subsection{Experiment Setting} \label{subsection:expsetting}
We evaluate the proposed approach on three video retrieval tasks, namely Near-Duplicate Video Retrieval (NDVR), Fine-grained Incident Video Retrieval (FIVR), and Event Video Retrieval (EVR). In all cases, we report the mean Average Precision (mAP).

\textbf{Training Dataset.}
We leverage the \textbf{VCDB}~\cite{jiang2014vcdb} dataset as the training dataset. The core dataset of VCDB has 528 query videos and 6,139 positive pairs, and the distractor dataset has 100,000 distractor videos, of which we successfully downloaded 99,181 of them.

\textbf{Evaluation Dataset.}
For models trained on the VCDB dataset, we test them on the \textbf{CC\_WEB\_VIDEO}~\cite{wu2007practical} dataset for NDVR task, \textbf{FIVR-200K} for FIVR task and \textbf{EVVE}~\cite{revaud2013event} for EVR task. For a quick comparison of the different variants, the \textbf{FIVR-5K} dataset as in~\cite{kordopatis2019visil} is also used. 
The CC\_WEB\_VIDEO dataset contains 24 query videos and 13,129 labeled videos;
The FIVR-200K dataset includes 225,960 videos and 100 queries, it consists of three different fine-grained video retrieval tasks: (1) Duplicate Scene Video Retrieval, (2) Complementary Scene Video Retrieval and (3) Incident Scene Video Retrieval;
The EVVE dataset is designed for the EVR task, it consists of 2,375 videos and 620 queries.

\textbf{Implementation Details.}
For feature extraction, we extract one frame per second for all videos. For all retrieval tasks, we extract the frame-level features following the scheme in Section~\ref{subsection:featextract}. The intermediate features are all extracted from the output of four residual blocks of ResNet-50~\cite{he2016deep}. PCA trained on 997,090 randomly sampled frame-level descriptors from VCDB is applied to both iMAC and $\text{L}_3$-iRMAC features to perform whitening and reduce its dimension from 3840 to 1024. Finally, $\ell_2$-normalization is applied.

For the Transformer model, it is implemented with one single layer, eight attention heads, dropout\_rate set to 0.5, and the dimension of the feed-forward layer set to 2048.

During training, all videos are padded to 64 frames (if longer, a random segment with a length of 64 is extracted), and the full video is used in the evaluation stage. Adam~\cite{kingma2014adam} is adopted as the optimizer, with the initial learning rate set to $10^{-5}$, and cosine annealing learning rate scheduler~\cite{loshchilov2016sgdr} is used. The model is trained with batch size 64 for 40 epochs, and $16\times64$ negative samples sampled from the distractors are sent to the memory bank each batch, with a single device with four Tesla-V100-SXM2-32GB GPUs, the size of the memory bank is equal to 4096.
The code is implemented with PyTorch~\cite{Paszke2019PyTorchAI}, and distributed training is implemented with Horovod~\cite{Sergeev2018HorovodFA}.

\subsection{Ablation Study} \label{subsection:ablation}

\textbf{Models for Temporal Context Aggregation.}
In Table~\ref{subtab:model}, we compare the Transformer with prior temporal context aggregation approaches, \ie, NetVLAD~\cite{arandjelovic2016netvlad}, LSTM~\cite{hochreiter1997long} and GRU~\cite{cho2014properties}. All models are trained on VCDB dataset with iMAC feature and evaluated on all three tasks of FIVR-5K, and dot product is used for similarity calculation for both train and evaluation. The classic recurrent models (LSTM, GRU) do not show advantage against NetVLAD. However, with the help of self-attention mechanism, the Transformer model demonstrate excellent performance gain in almost all tasks, indicating its strong ability of long-term temporal dependency modeling.

\textbf{Frame Feature Representation.}
We evaluate the iMAC and $\text{L}_3$-iRMAC feature on the FIVR-200K dataset with cosine similarity, as shown in Table~\ref{subtab:feature}. With more local spatial information leveraged, $\text{L}_3$-iRMAC show consistent improvement against iMAC. 

\textbf{Loss function for contrastive learning.}
We present the comparison of loss functions for contrastive learning in Table~\ref{subtab:loss}. The InfoNCE loss show notable inferiority compared with Circle with default parameters $\tau=0.07,\gamma=256,m=0.25$. By adjusting the sensitive temperature parameter $\tau$ (set to $1/256$, equivalent with $\gamma=256$ in Circle loss), it still shows around 0.5\% less mAP.

\textbf{Size of the Memory Bank.}
In Table~\ref{subtab:banksz}, we present the comparison of different sizes of the memory bank. It is observed that a larger memory bank convey consistent performance gain, indicating the efficiency of utilizing large quantities of negative samples.
Besides, we compare our approach against the commonly used triplet based approach with hard negative mining~\cite{kordopatis2017dml} (without bank). The training process of the triplet-based scheme is extremely time-consuming (5 epochs, 5 hours on 32 GPUs), yet still show around 10\% lower mAP compared with the baseline (40 epochs, 15 minutes on 4 GPUs), indicating that compared with learning from hard negatives, to utilize a large number of randomly sampled negative samples is not only more efficient, but also more effective. 

\textbf{Momentum Parameter.}
In Table~\ref{subtab:momentum}, we present the ablation on momentum parameter of the modified MoCo~\cite{he2019momentum}-like approach, where a large queue is maintained to store the negative samples and the weight of the model is updated in a moving averaged manner.
We experimented with different momentum ranging from 0.1 to 0.999 (with queue length set to 65536), but none of them show better performance than the baseline approach as reported in Table~\ref{subtab:banksz}, we argue that the momentum mechanism is a compromise for larger memory. as the memory bank is big enough in our case, the momentum mechanism is not needed. 

\textbf{Similarity Measure.}
We evaluate the video-level features with cosine similarity, and frame-level features following the setting of ViSiL~\cite{kordopatis2019visil}, \ie, chamfer similarity, symmetric chamfer similarity, and chamfer similarity with similarity comparator (the weights are kept as provided by the authors). Table~\ref{subtab:simmeasure} presents the results on FIVR-5K dataset. Interestingly, the frame-level similarity calculation approach outperforms the video-level approach by a large margin, indicating that frame-level comparison is important for fine-grained similarity calculation between videos. Besides, the comparator network does not show as good results as reported, we argue that this may be due to the bias between features.

Next, we only consider the Transformer model trained with $\text{L}_3$-iRMAC feature and Circle loss in the following experiments, denoted as TCA (Temporal Context Encoding for Video Retrieval). With different similarity measures, all four approaches are denoted as $\text{TCA}_{c}$ (cosine), $\text{TCA}_{f}$ (chamfer), $\text{TCA}_{sym}$ (symmetric-chamfer), $\text{TCA}_{v}$ (video comparator) for simplicity.

\subsection{Comparison Against State-of-the-art} \label{paragraph:compsota}
\begin{figure*}[t]
    \centering
    \includegraphics[width=\textwidth]{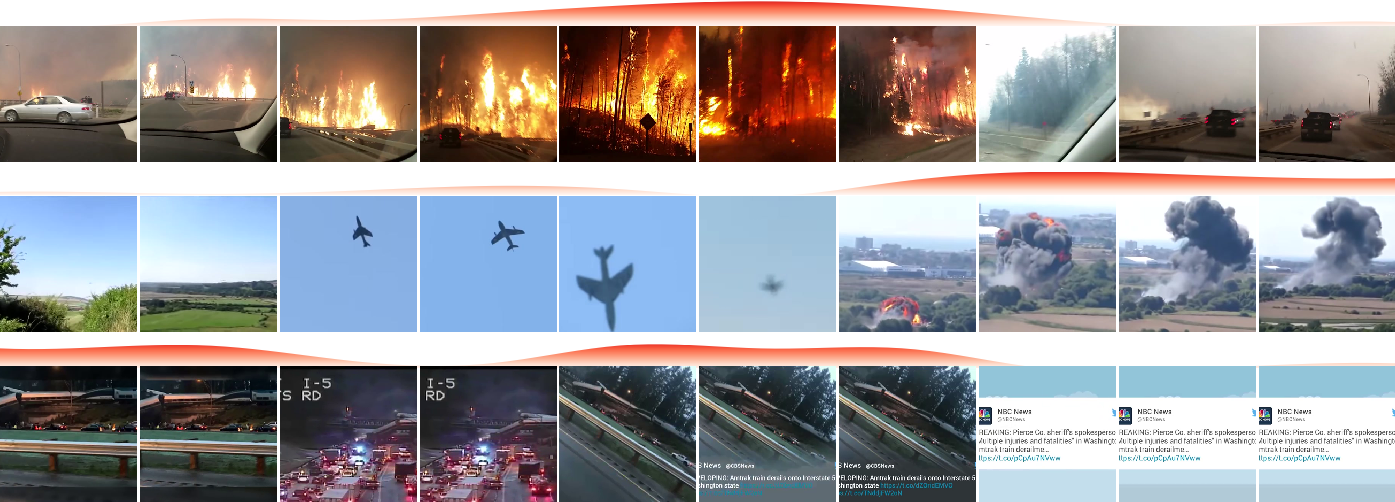}
    \caption{\textbf{Visualization of average attention weight (response) of example videos in FIVR.} The weights are normalized and interpolated for better visualization, and darker color indicates higher average response of the corresponding frame. Each case tends to focus on salient and informative frames: video \#1 focuses on key segments about the fire; video \#2 has a higher focus on the explosion segment; and video \#3 selectively ignores the meaningless ending.}
    \label{fig:attn}
\end{figure*}

\noindent\textbf{Near-duplicate Video Retrieval.} \label{paragraph:ndvr}
We ﬁrst compare TCA against state-of-the-art methods on several versions of CC\_WEB\_VIDEO~\cite{wu2007practical}. The benchmark approaches are Deep Metric Learning (DML)~\cite{kordopatis2017dml}, the Circulant Temporal Encoding (CTE)~\cite{revaud2013event}, and Fine-grained Spatio-Temporal Video Similarity Learning (ViSiL), we report the best results of the original paper. As listed in Table \ref{tab:ccweb}, we report state-of-the-art results on all tasks with video-level features, and competitive results against $\text{ViSiL}_v$ with refined frame-level features. To emphasize again, our target is to learn a good video representation, and the similarity calculation stage is expected to be as simple and efficient as possible, therefore, it is fairer to compare $\text{TCA}_{f}$ with $\text{ViSiL}_f$, as they hold akin similarity calculation approach. 
\begin{table}[htb]
\setlength{\tabcolsep}{3pt}
\caption{\textbf{mAP on 4 versions of CC\_WEB\_VIDEO.} Following the setting in ViSiL~\cite{kordopatis2019visil}, (*) denotes evaluation on the entire dataset, and subscript $c$ denotes using the cleaned version of the annotations.} \label{tab:ccweb}
{\small
\begin{tabular}{llcccc}
\toprule
\multicolumn{2}{c}{\multirow{2}{*}{Method}} & \multicolumn{4}{c}{CC\_WEB\_VIDEO}        \\
\cmidrule{3-6}
\multicolumn{2}{c}{}                        & cc\_web & cc\_web* & $\text{cc\_web}_c$ & $\text{cc\_web}_c$* \\
\midrule
Video-  & DML~\cite{kordopatis2017dml}         & 0.971   & 0.941    & 0.979    & 0.959     \\
level  & $\text{TCA}_c$       &\textbf{0.973}   & \textbf{0.947}    & \textbf{0.983}    & \textbf{0.965}     \\
\cmidrule{1-6}
                              & CTE~\cite{revaud2013event}         & \textbf{0.996}   & -        & -        & -         \\
                              & $\text{ViSiL}_f$~\cite{kordopatis2019visil}     & 0.984   & 0.969    & 0.993    & 0.987     \\
Frame-                        & $\text{ViSiL}_{sym}$~\cite{kordopatis2019visil}   & 0.982   & 0.969    & 0.991    & 0.988     \\
level                         & $\text{ViSiL}_v$~\cite{kordopatis2019visil}     & 0.985   & \textbf{0.971}    & \textbf{0.996}    & \textbf{0.993} \\
                              & $\text{TCA}_f$       & 0.983   & 0.969    & 0.994    & 0.990     \\
                              & $\text{TCA}_{sym}$   & 0.982    & 0.962    & 0.992    & 0.981     \\
\bottomrule
\end{tabular}
}
\end{table}

\noindent\textbf{Fine-grained Incident Video Retrieval.} \label{paragraph:fivr}
We evaluate TCA against state-of-the-art methods on FIVR-200K~\cite{kordopatis2019fivr}. We report the best results reported in the original paper of DML~\cite{kordopatis2017dml}, Hashing Codes (HC)~\cite{song2013effective}, ViSiL~\cite{kordopatis2019visil}, and their re-implemented DP~\cite{chou2015pattern} and TN~\cite{tan2009scalable}. As shown in Table~\ref{tab:fivr}, the proposed method shows a clear performance advantage over state-of-the-art methods with video-level features ($\text{TCA}_{c}$), and deliver competitive results with frame-level features ($\text{TCA}_{f}$).
Compared with $\text{ViSiL}_{f}$, we show a clear performance advantage even with a more compact frame-level feature and simpler frame-frame similarity measure.

A more comprehensive comparison on performance is given in Fig.~\ref{fig:perf}. The proposed approach achieves the best trade-off between performance and efficiency with both video-level and frame-level features against state-of-the-art methods.
When compared with $\text{ViSiL}_{v}$, we show competitive results with about 22x faster inference time. Interestingly, our method slightly outperforms $\text{ViSiL}_{v}$ in ISVR task, indicating that by conducting temporal context aggregation, our model might show an advantage in extracting semantic information.

\begin{figure}[htb]
\begin{subfigure}{.49\textwidth}
  \centering
  \includegraphics[width=1.0\linewidth]{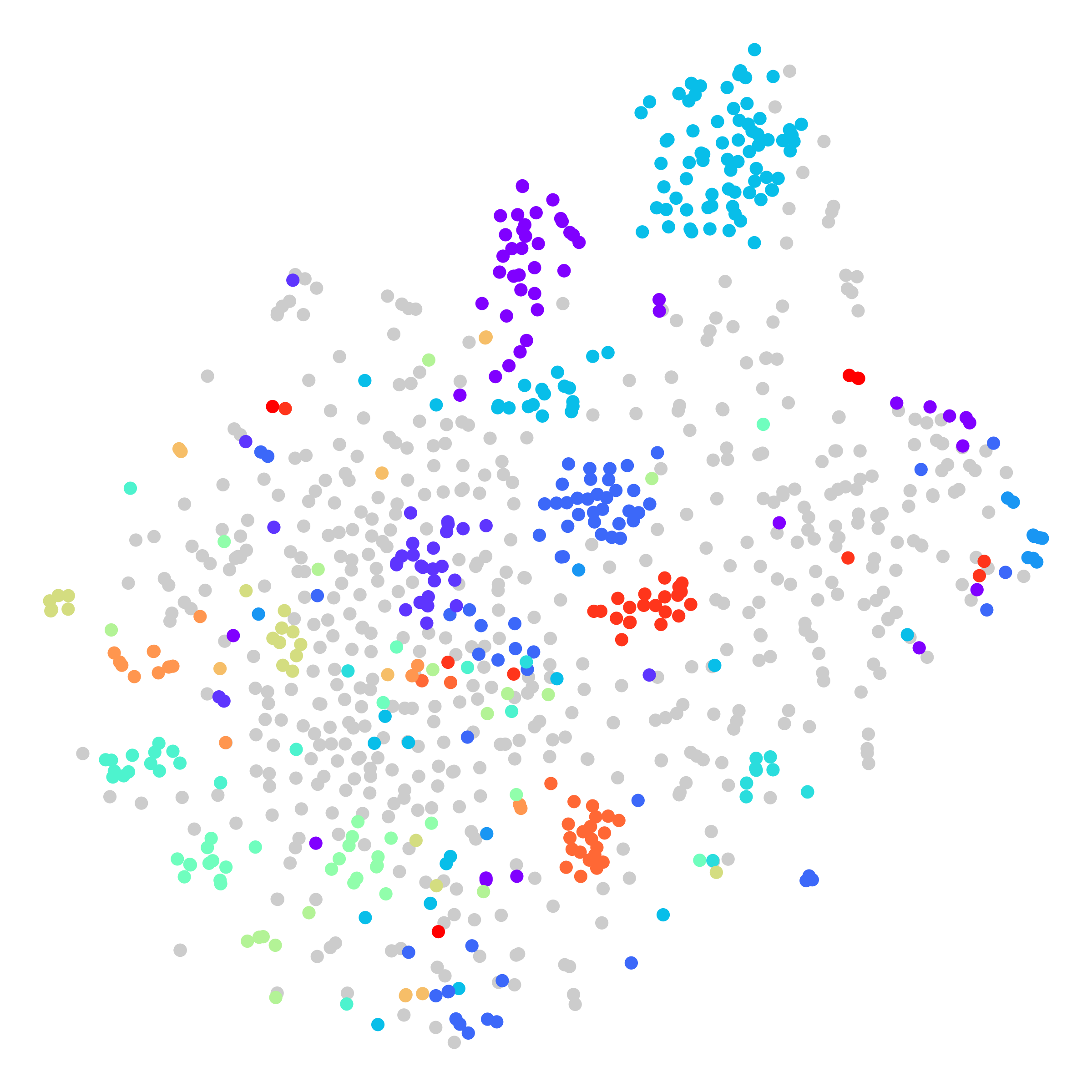}  
  \caption{DML~\cite{kordopatis2017dml}}
  \label{fig:dml}
\end{subfigure}
\begin{subfigure}{.49\textwidth}
  \centering
  \includegraphics[width=1.0\linewidth]{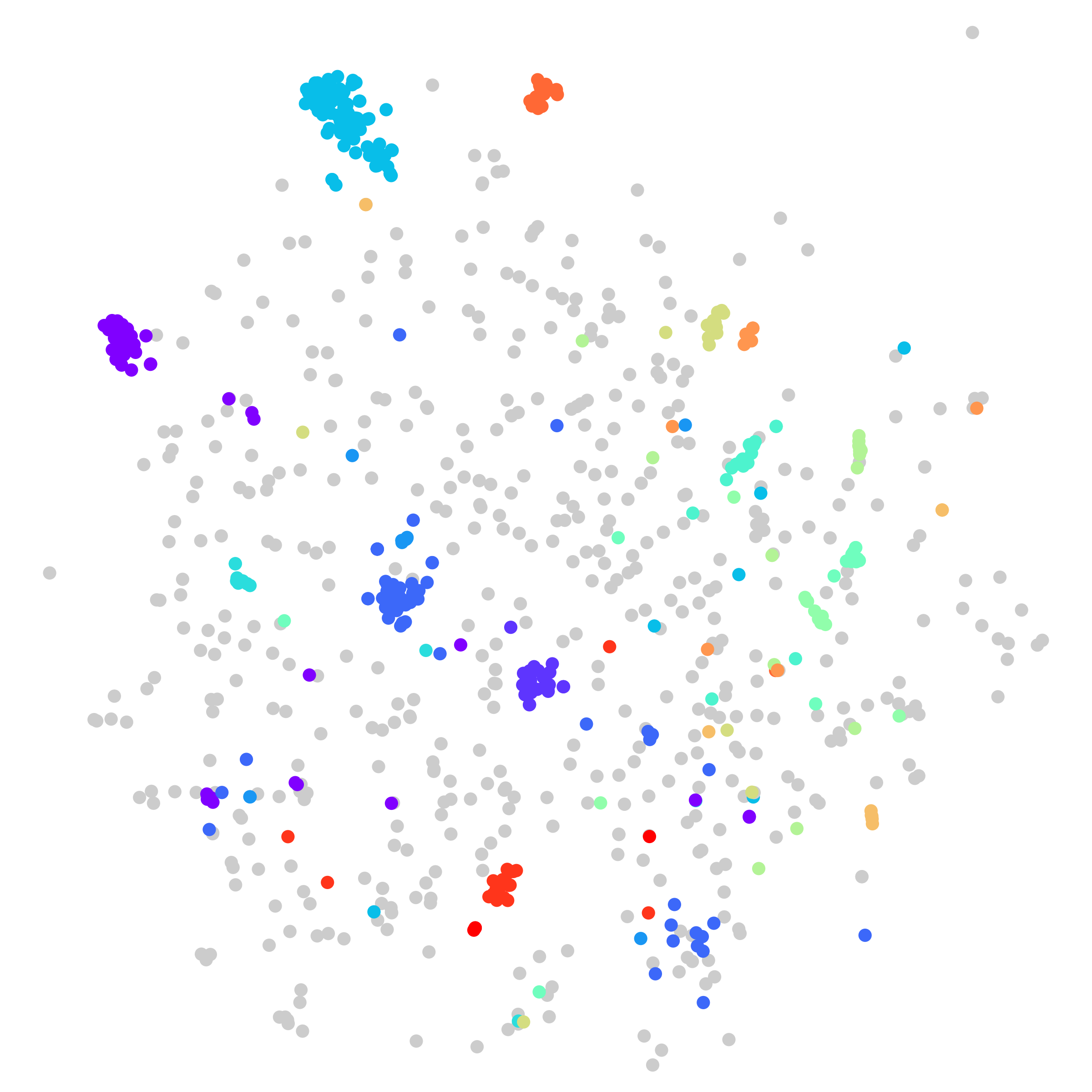}  
  \caption{Ours ($\text{TCA}_{c}$)}
  \label{fig:ours}
\end{subfigure}
\caption{\textbf{Visualization of video-level features on a subset of FIVR-5K with t-SNE.} Each color represents samples corresponding to one single query, and distractors are colored with faded gray. Both our method and DML are trained on VCDB~\cite{jiang2014vcdb} dataset. (\textit{Best viewed in color})}
\label{fig:tsne}
\end{figure}

\begin{table}[htb]
\setlength{\tabcolsep}{5.25pt}
{\small
\begin{tabular}{llcccc}
\toprule
\multicolumn{2}{c}{\multirow{2}{*}{Method}} & \multicolumn{3}{c}{FIVR-200K} & \multirow{2}{*}{EVVE} \\
\cmidrule{3-5}
\multicolumn{2}{c}{}                        & DSVR     & CSVR     & ISVR   \\
\midrule
                    & DML~\cite{kordopatis2017dml}                   & 0.398    & 0.378    & 0.309  & - \\
Video-              & HC~\cite{song2013effective}                    & 0.265    & 0.247    & 0.193  & - \\
level               & LAMV+QE~\cite{baraldi2018lamv}                 & -        & -        & -      & 0.587 \\
                    & $\text{TCA}_c$         & \textbf{0.570}   & \textbf{0.553}   & \textbf{0.473}    & \textbf{0.598} \\
\cmidrule{1-6}
                    & DP~\cite{chou2015pattern}                         & 0.775    & 0.740    & 0.632   & -     \\
                    & TN~\cite{tan2009scalable}                         & 0.724    & 0.699    & 0.589   & -     \\
                    & $\text{ViSiL}_f$~\cite{kordopatis2019visil}       & 0.843    & 0.797    & 0.660   & 0.597 \\
Frame-              & $\text{ViSiL}_{sym}$~\cite{kordopatis2019visil}   & 0.833    & 0.792    & 0.654   & 0.616 \\
level               & $\text{ViSiL}_v$~\cite{kordopatis2019visil}       & \textbf{0.892}      & \textbf{0.841}    & 0.702   & 0.623 \\
                    & $\text{TCA}_f$         & 0.877    & 0.830    & \textbf{0.703}    & 0.603 \\
                    & $\text{TCA}_{sym}$         & 0.728    & 0.698    & 0.592    & \textbf{0.630} \\
\bottomrule
\end{tabular}
}
\caption{\textbf{mAP on FIVR-200K and EVVE.} The proposed approach achieves the best trade-off between performance and efficiency with both video-level and frame-level features against state-of-the-art methods.} \label{tab:fivr}
\end{table}

\noindent\textbf{Event Video Retrieval.} \label{paragraph:evr}
For EVR, we compare TCA with Learning to Align and Match Videos (LAMV)~\cite{baraldi2018lamv} with Average Query Expansion (AQE)~\cite{douze2013stable} and ViSiL~\cite{kordopatis2019visil} on EVVE~\cite{revaud2013event}. We report the results of LAMV from the original paper, and the re-evaluated ViSiL (the reported results are evaluated on incomplete data). As shown in Table \ref{tab:fivr}, $\text{TCA}_{sym}$ achieves the best result. Surprisingly, our video-level feature version $\text{TCA}_{c}$ also report notable results, this may indicate that the temporal information and fine-grained spatial information are not necessary for event video retrieval task.

\subsection{Qualitative Results}
We demonstrate the distribution of video-level features on a randomly sampled subset of FIVR-5K with t-SNE~\cite{maaten2008visualizing} in Fig.~\ref{fig:tsne}. Compared with DML, the clusters formed by relevant videos in the refined feature space obtained by our approach are more compact, and the distractors are better separated;
To better understand the effect of the self-attention mechanism, we visualize the average attention weight (response) of three example videos in Fig.~\ref{fig:attn}. The self-attention mechanism helps expand the vision of the model from separate frames or clips to almost the whole video, and conveys better modeling of long-range semantic dependencies within the video. As a result, informative frames describing key moments of the event get higher response, and the redundant frames are suppressed.

\section{Conclusion}
In this paper, we present TCA, a video representation learning network that incorporates temporal-information between frame-level features using self-attention mechanism to help model long-range semantic dependencies for video retrieval. To train it on video retrieval datasets, we propose a supervised contrastive learning method. With the help of a shared memory bank, large quantities of negative samples  are utilized efficiently with no need for manual hard-negative sampling. Furthermore, by conducting gradient analysis, we show that our proposed method has the property of automatic hard-negative mining which could greatly improve the final model performance. Extensive experiments are conducted on multi video retrieval tasks, and the proposed method achieves the best trade-off between performance and efficiency with both video-level and frame-level features against state-of-the-art methods.

{\small
\bibliographystyle{ieee_fullname}
\bibliography{main_arxiv}
}

\end{document}